\documentclass{article}

\usepackage{microtype}
\usepackage{graphicx}
\usepackage{subfigure}
\usepackage{booktabs} 

\usepackage{hyperref}

\usepackage[accepted]{icml2024}

\usepackage{amsmath}
\usepackage{amssymb}
\usepackage{mathtools}
\usepackage{amsthm}

\usepackage[capitalize,noabbrev]{cleveref}

\theoremstyle{plain}

\theoremstyle{definition}

\theoremstyle{remark}

\usepackage[textsize=tiny]{todonotes}

\usepackage[all]{hypcap}
 \creflabelformat{equation}{#1#2#3}
\DeclareRobustCommand{\parhead}[1]{\textbf{#1}~}
\usepackage{caption}
\usepackage{subcaption}

\newcommand{\E}{\mathbb{E}}

\icmltitlerunning{Do Large Language Models Perform the Way People Expect?}

\begin{document}

\twocolumn[
\icmltitle{Do Large Language Models Perform the Way People Expect?\\ Measuring the Human Generalization Function}

\icmlsetsymbol{equal}{*}

\begin{icmlauthorlist}
\icmlauthor{Keyon Vafa}{harvard}
\icmlauthor{Ashesh Rambachan}{mit}
\icmlauthor{Sendhil Mullainathan}{chicago}
\end{icmlauthorlist}

\icmlaffiliation{harvard}{Harvard University}
\icmlaffiliation{mit}{Massachusetts Institute of Technology}
\icmlaffiliation{chicago}{University of Chicago Booth}

\icmlcorrespondingauthor{Keyon Vafa}{kvafa@g.harvard.edu}

\icmlkeywords{Machine Learning, ICML}

\vskip 0.3in
]

\printAffiliationsAndNotice{}  

\begin{abstract}
What makes large language models (LLMs) impressive is also what makes them hard to evaluate: their diversity of uses. To evaluate these models, we must understand the purposes they will be used for. We consider a setting where these deployment decisions are made by people, and in particular, people's \textit{beliefs} about where an LLM will perform well. We model such beliefs as the consequence of a human generalization function: having seen what an LLM gets right or wrong, people generalize to where else it might succeed. We collect a dataset of 19K examples of how humans make generalizations across 79 tasks from the MMLU and BIG-Bench benchmarks. We show that the human generalization function can be predicted using NLP methods: people have consistent structured ways to generalize. We then evaluate LLM alignment with the human generalization function. Our results show that --- especially for cases where the cost of mistakes is high --- more capable models (e.g. GPT-4) can do worse on the instances people choose to use them for,  exactly because they are not aligned with the human generalization function. 
~\looseness=-1
\end{abstract}

\section{Introduction}
Large language models (LLMs) afford a remarkable diversity of uses. This diversity offers immense promise: the same model can be used to help software engineers write code and to summarize a doctor's notes from a clinical appointment. However, this same diversity poses an evaluation problem: how should we evaluate a model that seems capable of doing many things?

Evaluating LLMs in the same way as supervised learning models --- by prespecifying a task and evaluating on a relevant benchmark --- undersells the capabilities of LLMs: 
LLMs are capable of performing many tasks, not all of which can be enumerated. 
Moreover, many of the tasks that LLMs will be deployed to perform do not have existing benchmarks. For example, if a business owner wants to use an LLM to respond to client emails, which benchmark dataset should they use to evaluate LLMs? Creating a new benchmark for each possible task is infeasible.

Crucially, in many instances, the decisions about where an LLM will be deployed are made by people. 
These decisions are often driven by where they believe a model will perform well \citep{lubars2019ask,lai2022human}. 
Assessing the real-world performance of LLMs therefore 
requires understanding how people form beliefs about their capabilities.

In this paper, we introduce a framework for quantifying people's beliefs about LLM capabilities.
In our framework, these beliefs are determined through interaction 
and evolve via a \textit{human generalization function}:
humans ask questions, observe how an LLM responds, and make inferences about how it would respond to other questions. 
This is an act of human \text{generalization}, similar to how humans judge the expertise of other people based on prior interactions. 
For example, people may expect that a model that can answer college physics questions is capable of answering elementary math questions, but make no inferences about its ability to answer questions about Japanese literature. 
~\looseness=-1

\begin{figure*}
  \centering
  \makebox[\textwidth][c]{\includegraphics[width=\textwidth]{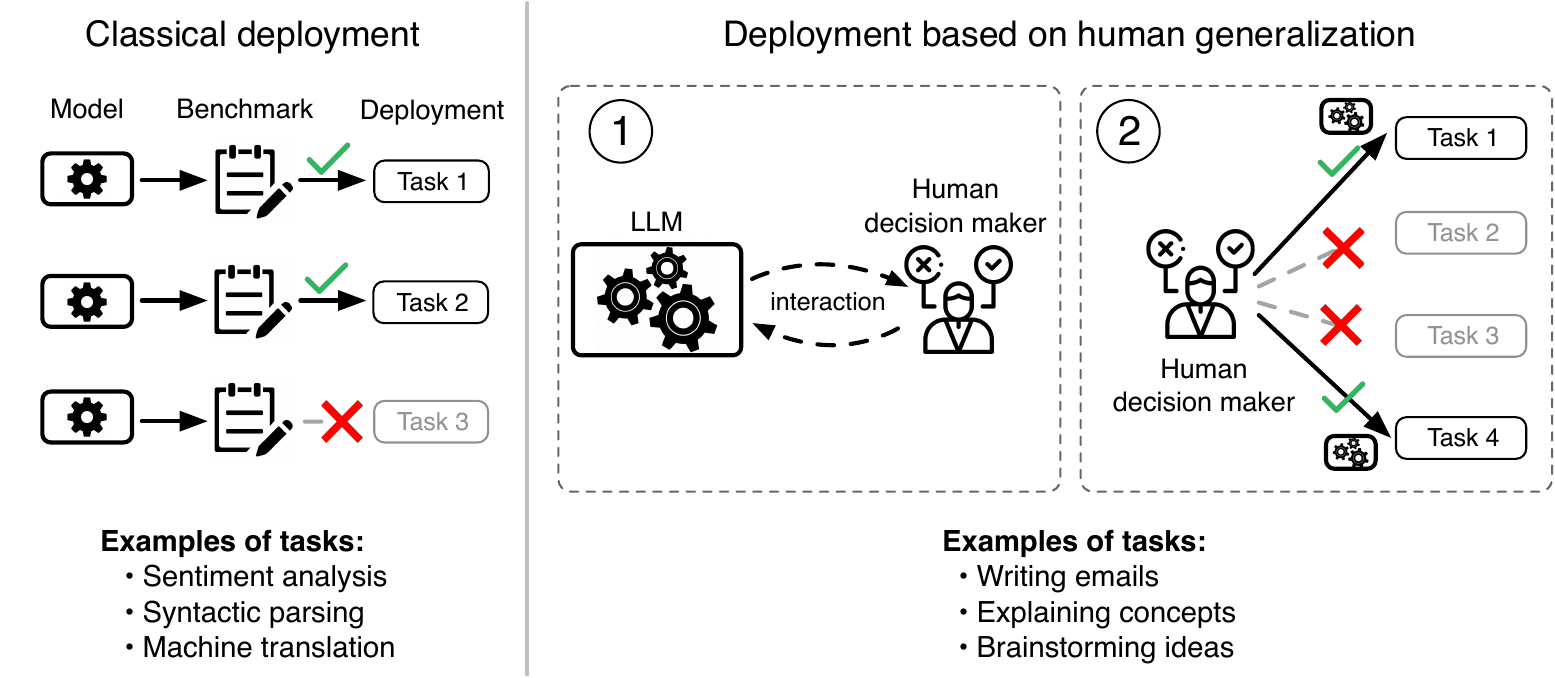}}
  \caption{Classically, ML models are deployed to perform tasks based on benchmark performance (left). When deployment is based on human generalization (right), a human decision maker first interacts with a model to assess its capabilities, and then the model is deployed to perform tasks the decision maker believes it will perform well on. The model's deployed performance depends on how well aligned its capabilities are with the human generalization function.}
  \label{fig:overview}
\end{figure*}

We propose an alignment problem based on the human generalization function: the best LLM is the one that allows humans to make the most reliable inferences about where it will succeed. We show that misalignment with the human generalization function can have harmful consequences for deployment: 
a misaligned model that outperforms another model on all individual questions can be less effective if a human deploys it to answer questions it is not capable of answering. 
~\looseness=-1

To study the human generalization function empirically, we collect data about the generalizations people make about LLM capabilities.
Specifically, we show humans how an LLM responded to one question, and ask them whether it affects their beliefs about how the same model would respond to another question. 
We show this function is sparse: most pairs of questions do not result in people updating their beliefs. Therefore, we use a bandit approach to target pairs of questions where humans update their beliefs. In total, we collect 18,972 examples of human generalizations involving pairs of questions from 79 distinct LLM evaluation tasks. 
~\looseness=-1

We then pose a new task: how well can the questions that change human beliefs be predicted? 
We show that belief changes can be predicted using NLP methods. 
BERT outperforms larger and more recent language models, suggesting that efficacy in predicting human generalization does not strictly scale with size or complexity.
We release all survey data in hopes that it will foster further research in modeling the human generalization function.\footnote{\url{https://github.com/keyonvafa/human-generalization-llms}}

Finally, we use our model of human generalizations to evaluate how aligned LLMs are with the human generalization function. 
In settings where the cost of mistakes is low, alignment increases with model size. 
However, in settings where the cost of mistakes is high, the trend reverses: based on short interactions, humans reach overconfident conclusions about the capabilities of the largest models. 

The rest of the paper is structured as follows: \Cref{sec:framework} introduces our framework. \Cref{sec:survey} describes how we collect data about human generalizations. \Cref{sec:predicting_belief_changes} shows that the human generalization function can be predicted. Finally, \Cref{sec:evaluating_alignment} evaluates the alignment between LLMs and the human generalization function.

\section{Framework}
\label{sec:framework}

Consider a setting in which a large language model (LLM) is deployed to answer questions from a distribution of questions chosen by a human. 
We show that traditional evaluation techniques, which assume the distribution of questions to be fixed, can provide misleading measures of deployed model performance. Instead, evaluation requires modeling how humans form deployment distributions. To study this process, we focus on the case where a human deploys an LLM to answer questions they believe it can answer correctly. Their beliefs are formed by generalization: humans assess an LLM's overall capabilities from a small set of interactions. 

\parhead{Notation.}
We use $\Sigma^*$ to refer to the set of all strings from an alphabet $\Sigma$. We define a set of questions $X \subseteq \Sigma^*$ to be a subset of the set of all strings. We define a model $f: X \to \Sigma^*$ to take as input a question and return a string from the alphabet.\footnote{Our framework extends naturally to the case in which a model is random, returning a probability distribution $\Delta(\Sigma^*)$ over strings in response to any question $x \in X$.}
Additionally, $t: X \times \Sigma^* \to \{0, 1\}$ is a function that assesses the quality of a response to a question, where $t(x, y) = 1$ indicates that $y \in \Sigma^*$ is a correct response to question $x \in X$ and $t(x, y) = 0$ indicates an incorrect response. (The quality of responses could also be real-valued but we use a binary function without loss of generality.) 

\parhead{Evaluation when people choose questions.}
Evaluating machine learning models involves making assumptions about how they will be deployed. For classical supervised learning methods, these assumptions are often straightforward: they will be deployed to answer the kinds of questions they were trained to answer. Therefore, models are typically evaluated against a \textit{fixed deployment distribution} $p(x)$, where overall effectiveness is
\begin{equation}
    \label{eqn:fixed_distribution_evaluation}
    \textstyle \sum_x p(x) t(x, f(x)).
\end{equation}
For example, a model developed solely to perform sentiment analysis would be evaluated based on its performance on a sentiment analysis benchmark.\footnote{Even this standard evaluation rests on the assumption that deployment will resemble the benchmark dataset. When this is not the case, such as under distribution shift, standard evaluation could be a misleading indicator of real-world performance \citep{koh2021wilds}.}

General-purpose models like LLMs differ from specific-purpose methods in that there is no fixed set of tasks for which they will be deployed; the capabilities of LLMs far exceed the kinds of questions they were trained to answer. 
Moreover, many tasks LLMs may be deployed for (e.g. writing emails) do not have the well-defined benchmark datasets that are required for classical evaluation.  

Instead, we focus on a setting where humans choose the questions a model is deployed to answer \citep{lubars2019ask,lai2022human}. 
Humans deploy an LLM to answer questions according to a \textit{human deployment distribution} $h(x|f)$, a probability distribution over inputs that may depend on the  model $f(\cdot)$. The human-deployed performance of a model is 
\begin{equation}
    \label{eqn:deployed_performance}
    \textstyle \sum_x h(x|f)t(x, f(x)).
\end{equation}

Human-deployed performance depends on a model in two ways. 
The first is direct and well-studied: an LLM that answers more questions right will have higher values of $t(x, f(x))$. But there is a more subtle second effect: the chosen questions $h(x|f)$ may depend on the model. If a human deploys a model to questions it is not capable of answering, its deployment performance will suffer. 
This second effect is omitted when a model is evaluated using a fixed deployment distribution as in \Cref{eqn:fixed_distribution_evaluation}. 

That the distribution over deployment questions depends on the model itself has implications for model comparison. 
Consider two models $f_1(\cdot), f_2(\cdot)$. 
We say $f_1(\cdot)$ \textit{dominates} $f_2(\cdot)$ if every question that $f_2(\cdot)$ answers correctly is also answered correctly by $f_1(\cdot)$:
\begin{equation}
    \label{eqn:weak_dominance}
    \{ x \colon t(x, f_2(x)) = 1 \} \subseteq \{ x \colon t(x, f_1(x)) = 1 \}.
\end{equation}
In this case, we write $f_1(\cdot) \succeq f_2(\cdot)$. 
A model that dominates another model will have performance at least as good as that model when evaluated using a fixed deployment distribution (\Cref{eqn:fixed_distribution_evaluation}). That is, 
if $f_1(\cdot) \succeq f_2(\cdot)$,
\begin{equation}
\label{eqn:dominance_fixed_deployment}
    \textstyle \sum_x p(x)t(x, f_1(x))  \geq \sum_x p(x) t(x, f_2(x)),
\end{equation}
for all fixed deployment distributions $p(x)$. However, when deployment distributions are dictated by humans, the questions an LLM is chosen to answer can depend on the model and the relationship may flip:  
\begin{equation}
    \label{eqn:dominance_human_deployent}
    \textstyle \sum_x h(x|f_1)t(x, f_1(x)) < \sum_x   h(x|f_2)t(x, f_2(x))
\end{equation}
In \Cref{app:human_distribution_dominance}, we show that human deployment does not preserve dominance orderings of models. 
Consider any pair of ``non-trivial'' models $f_1(\cdot), f_2(\cdot)$ (i.e., neither model answers all questions correctly or incorrectly) satisfying $f_1(\cdot) \succeq f_2(\cdot)$. There exists some human deployment distribution $h(x | f)$ under which the human-deployed performance of $f_2(\cdot)$ strictly improves upon that of $f_1(\cdot)$.
Evaluating the deployed performance of a model therefore requires incorporating the human deployment distribution $h(x|f)$. 

\parhead{How people choose questions.}
So far, we have left the human deployment distribution $h(x | f)$ unmodeled. 
We next consider one possible mechanism that would give rise to human deployment distributions: humans choose deployment questions based on their beliefs about a model's ability. 
For example, a business owner may decide an LLM is effective at summarizing meeting notes but not at responding to emails, and delegate responsibilities as such. 
~\looseness=-1

Specifically, let $0 \leq b(x|f) \leq 1$ denote a human's belief that a model $f(\cdot)$ will respond to input $x$ correctly. 
We assume a human chooses a deployment distribution based on their beliefs about the model's capabilities: $h(x|f) = h(b(x|f))$ for some transformation $h(\cdot)$. This transformation can take many forms;
for example, humans may deploy a model to answer all questions $x$ where their belief of success is above some threshold, $b(x|f) > \tau$.  

How do humans form beliefs about the capabilities of large language models? 
Beliefs may develop through interaction. It is infeasible for a human to assess an LLM by asking it every possible question. Instead, users \textit{generalize}: given a model, humans ask questions, observe how the model responds, and draw conclusions about how the model would respond to other questions. When a human thinks a question is directly relevant to another, they may update their beliefs; for example, they may assume that an LLM that answers a simple addition question correctly would also be able to answer a similarly worded one correctly. On the other hand, if two questions are unrelated, they may not update their beliefs at all.
~\looseness=-1

More formally, for two questions $x$ and $x^\prime$, we model this updating through the \textit{human generalization function}, $b(x \mid x^\prime, f)$, 
which summarizes a human's beliefs about an LLM's correctness on question $x$ after observing question $x^\prime$ and the model's response $f(x^\prime)$.
Such a human generalization function may arise, for example, if humans have some joint beliefs about the LLM's correctness on questions $x$ and $x^\prime$ and update their beliefs according to Bayes rule. 
Humans may generalize about the performance of LLMs in many ways, and so we write the human generalization function to capture this phenomenon. 

Human generalizations dictate the specific questions a model will be put to answer. The deployed performance of a model (\Cref{eqn:deployed_performance}) therefore depends on how \textit{aligned} the model is with human generalizations. If, after a few interactions, a human can assess the questions a model is equipped to answer, we say that the model is aligned with the human generalization function and it will thus be deployed to answer questions it is capable of answering. On the other hand, if it is misaligned, humans may be overconfident or underconfident and deploy an LLM to answer a suboptimal set of questions. \Cref{fig:overview} contains an overview of our framework. 

As a concrete example, consider two LLMs $f_1(\cdot)$ and $f_2(\cdot)$ that can both answer every arithmetic question correctly. Assume that $f_1(\cdot)$ cannot answer any other question correctly, while $f_2(\cdot)$ can also answer questions about multivariable calculus correctly but not questions about single-variable calculus. $f_2(\cdot)$ dominates $f_1(\cdot)$, $f_2(\cdot) \succeq f_1(\cdot)$. 
If, after interacting with each model, a user correctly determines $f_1(\cdot)$ is only capable of answering basic arithmetic questions correctly, they may deploy it only to answer basic arithmetic questions. 
If they incorrectly determine that $f_2(\cdot)$ is capable of answering \textit{all} math questions correctly, the user may deploy it to answer all arithmetic, single-variable calculus, and multi-variable calculus questions. Despite $f_2(\cdot)$ dominating $f_1(\cdot)$ on every question, its average deployed error is worse because its capabilities are misaligned with the human generalization function. 
~\looseness=-1

The remainder of the paper is dedicated to empirically investigating human generalizations of LLMs.  
\Cref{sec:survey} describes the collection of a dataset of 19K human generalizations. 
In \Cref{sec:predicting_belief_changes}, we analyze this dataset and show that human beliefs evolve in predictable ways, and propose a new benchmark: predict how human beliefs will change from specific interactions. Finally, in \Cref{sec:evaluating_alignment}, we assess how aligned various LLMs are with the human generalizations we collect.

\begin{figure*}
  \centering
  \makebox[\textwidth][c]{\includegraphics[width=\textwidth]{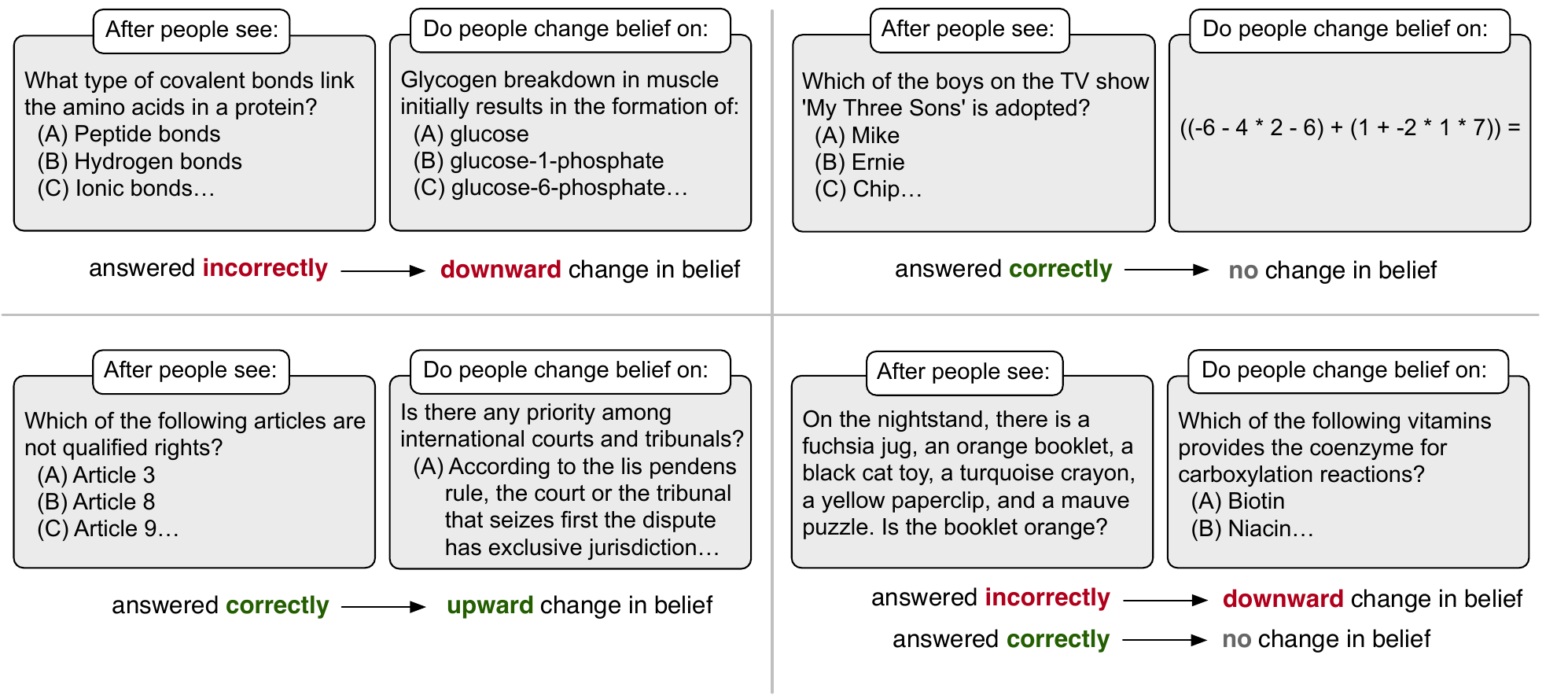}}
  \caption{Qualitative examples about question pairs and predicted belief changes.}
  \label{fig:generalization_examples}
\end{figure*}

\section{Collecting Data on Human Generalizations}
\label{sec:survey}
We conduct a survey to collect a dataset of 18,972 examples of human generalizations.
In this section, we detail the survey methodology and show that human generalizations are sparse: for most pairs of questions, humans do not update their beliefs for how a model will respond to one question after seeing how it responded to the other. 
We describe a bandit approach for finding pairs of questions that lead to belief changes, and upweight these pairs in our data collection. 
~\looseness=-1

\parhead{Survey design.}
We construct each example in our survey as follows: a human is first asked to predict how likely a large language model is to answer a given question $x$ correctly. 
They are then shown how an LLM responded to another question $x'$, and are asked to update their initial prediction. 
This design elicits a person's \textit{prior} belief of correctness $b(x|f)$, the outcome of their generalization function $b(x|x', f)$, and the resulting change in beliefs: 
\begin{equation}
    \label{eqn:update}
    \Delta(x|x', f) = b(x|x', f) - b(x|f).
\end{equation}
We refer to the output of the human generalization function $b(x \mid x^\prime, f)$ as a \textit{posterior} belief, 
since it describes the human's beliefs after observing the LLM's response on another question. 
Since the survey questions are about hypothetical LLMs, we do not name specific LLMs when eliciting prior or posterior beliefs. 
See \Cref{fig:generalization_examples} for examples of question pairs in the survey.
 
We aim to measure human generalization across a wide variety of questions, while also requiring that questions have a single correct answer that can be verified automatically. 
Therefore, we base data collection on questions from widely used LLM benchmarks. 
Our dataset consists of questions from the 57 tasks in the Massive Multitask Language Understanding (MMLU) benchmark \citep{hendrycks2020measuring} and 22 tasks from the BIG-Bench Hard (BBH) benchmark \citep{suzgun2022challenging}, totaling 16,347 individual questions. 
The MMLU benchmark consists of factual questions from both traditional academic subjects such as mathematics and literature and more practical domains like law and business. 
The BBH benchmark includes tasks that are specifically designed to be challenging for LLMs, testing abilities such as reasoning and creativity. 
The diversity of questions in these datasets ensures that the survey captures a wide range of perceived capabilities. 
We include more details about these datasets and their construction in \Cref{app:dataset_construction}. 

We conduct all surveys on Prolific \citep{palan2018prolific}.
To take our surveys, respondents must first pass two comprehension checks to ensure their comprehension of our design. We collect 15 pairs of predictions from each respondent. We pay each respondent \$2.50, and the median survey completion time is 12 minutes, for an implied rate of \$12.50/hour. We received an IRB review and exemption for this study.

\begin{figure*}
  \centering
  \makebox[\textwidth][c]{\includegraphics[width=1.0\textwidth]{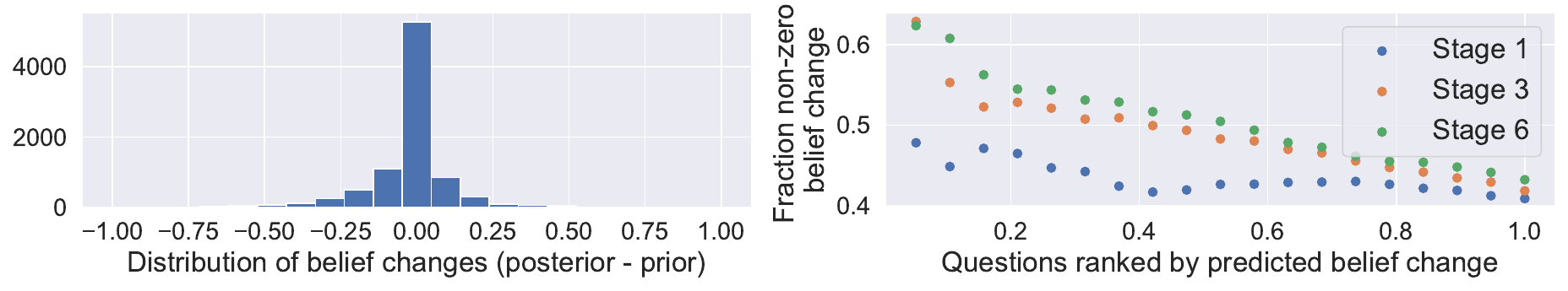}}
  \caption{Left: The human generalization function is sparse. Most pairs of randomly sampled questions result in no belief change. Right: The bandit is effective at identifying instances where beliefs change. The x-axis ranks bandit predictions by likelihood of belief change, while the y-axis shows the fraction actually containing belief changes (using a held-out set).
  Over time, the bandit becomes effective at finding non-zero belief changes.
  }
  \label{fig:sparsity_and_bandit}
\end{figure*}

\parhead{Bandit for finding non-sparse examples.}
We first conducted a pilot survey, where we randomly sampled pairs of questions from our dataset and asked users for their prior and posterior beliefs. Based on this pilot, it was clear that the human generalization function is \textit{sparse}: for most randomly selected pairs of questions, a human's prior beliefs are the same as their posterior beliefs (see \Cref{fig:sparsity_and_bandit}).

To focus on question-pairs that lead to meaningful human generalization, we modify the survey to encourage examples that will result in belief changes using a bandit approach. 
We divide the survey into seven stages.
After each stage, we train BERT \citep{devlin2018bert} to predict the pairs of questions that are likeliest to result in belief changes (more model details are in \Cref{sec:predicting_belief_changes}).
For the next stage, we upweight question pairs that are predicted to result in the large beliefs changes based on epsilon-greedy sampling \citep{kuleshov2014algorithms}, sampling the majority of questions from the top 10th percentile of the model's predicted likelihood change and the remaining questions from elsewhere in the distribution.
This is effective at finding question pairs with non-zero belief changes. \Cref{fig:sparsity_and_bandit} shows how the sampling approach becomes more successful over time. As the survey progresses, the model's predictions of the likeliest questions to result in belief changes improve.

After seven stages of survey collection, we collected a total of 18,480 response pairs. We performed one final stage of data collection to use as a test set, collecting at least 8 responses for each of 492 response pairs to be used for model evaluation. We aggregated belief changes by taking the majority response for each pair. 
For this final stage, we targeted a 50/50 balance between pairs of questions that change beliefs and those that do not. We did this by sampling 2/3 of the dataset from the top 1-percent of the bandit's distribution and the remaining 1/3 from the bottom 1-percent. This test dataset consists of 492 response pairs, 43\% of which resulted in a belief change. 
~\looseness=-1

\section{Modeling Human Generalizations}
\label{sec:predicting_belief_changes}

Having collected data, we turn to modeling human generalizations: how predictable are human belief changes from the text of prior interactions? 
We consider different models for predicting these changes, which we will use in \Cref{sec:evaluating_alignment} to evaluate LLM alignment with human generalizations. 

\parhead{Predicting belief changes.}
Recall that an individual's belief change about how an LLM $f(\cdot)$ will respond to question $x$ given its response to $x'$ is given by $\Delta(x|x', f)$ (\Cref{eqn:update}). 
We define a benchmark task to predict whether an individual's belief will change: here, the goal is predicting the binary outcome $1(\Delta(x|x', f) \neq 0)$. We consider different models denoted by $g_\theta(x|x', f(x'))$, where each model predicts the likelihood of belief change.
We consider the following models:
~\looseness=-1
\begin{itemize}
    \item \textbf{Previous correct:} A baseline model that doesn't use text. It predicts whether a human's belief about question $x$ will change only from whether an LLM answered question $x'$ correctly. $g_\theta(x|x', f(x'))$ is modeled as a logistic regression.
    \item \textbf{Previous correct + same task:} Another baseline model that does not use text. This model adds one additional feature to the previous baseline: whether $x$ and $x'$ are from the same task (of the 79 possible tasks).
    The model is trained as a logistic regression.
    \item \textbf{Fixed embeddings + XGBoost:} This model embeds each question using fixed sentence embeddings from \href{https://www.voyageai.com}{VoyageAI}. Embeddings for $x'$ and $x$ are concatenated, and XGBoost \citep{chen2016xgboost} is used to predict belief change on $x$.
    \item \textbf{BERT:} An ensemble of 5 BERT-base models \citep{devlin2018bert} are fine-tuned, taking as input the pair of concatenated inputs $x'$ and $x$ along with a special token denoting whether $x'$ was answered correctly. Predictions are averaged across ensembles. 
    \item \textbf{Llama-2 7B/13B:} Llama-2 models \citep{touvron2023llama} are fine-tuned to predict belief change on $x$ using a custom prompt containing $x$, $x'$, and whether $x'$ was answered correctly. 
    \item \textbf{GPT-3.5 turbo (10-shot, exact inference):}  We prompt GPT-3.5 turbo \citep{brown2020language} using 10 other examples of generalizations and record the likelihood the model predicts `0' or `1'.
    \item \textbf{GPT-4 (10-shot, MC inference):} We prompt GPT-4 \citep{achiam2023gpt} using 10 other examples of generalizations. Since the GPT-4 API doesn't provide predicted probabilities, we sample 5 answers with temperature 1.0 and record the predicted probability as the Monte Carlo average.
\end{itemize}
Each model is trained on the initial 18,480 response pairs and evaluated on the test set of 492 aggregated labels. More details about models and evaluation are provided in \Cref{app:experimental_details}.

\begin{table*}
  \small
  \centering
  \begin{tabular} {l c c c c c c}
  \multicolumn{1}{c}{} & \multicolumn{3}{c}{\textbf{NLL}} & \multicolumn{3}{c}{\textbf{AUC}} \\  
  \toprule
  & Overall & Prev. correct & Prev. incorrect & Overall & Prev. correct & Prev. incorrect \\
    \midrule
  Previous correct & 0.666 & 0.638 & 0.694 & 0.599 & 0.500 & 0.500 \\
  Same subject + previous correct & 0.637 & 0.598 & 0.675 & 0.688 & 0.680 & 0.556 \\
  Fixed embeddings + XGBoost & 0.629 & 0.589 & 0.669 & 0.682 & 0.627 & 0.690 \\
  BERT (fine-tuned) & \textbf{0.593} & \textbf{0.545} & \textbf{0.640} &  \textbf{0.812} & 0.855 & 
  \textbf{0.775} \\ 
  Llama-2 (7B) (fine-tuned) & 0.652 & 0.620 & 0.690 & 0.755 & 0.827 & 0.668  \\
  Llama-2 (13B) (fine-tuned) & 0.624 & 0.591 & 0.663 &  0.787 & \textbf{0.867} & 0.772 \\
  GPT-3.5 (turbo) (10-shot, exact inference) & 0.912 & 0.911 & 1.094 & 0.537 & 0.528 & 0.418 \\
  GPT-4 (10-shot, MC inference) & 0.789 & 0.705 & 0.873 & 0.640 & 0.585 & 0.566  \\
  \bottomrule
  \end{tabular}
  \caption{Each model's performance at predicting human belief change on question $x$ after seeing an LLM's response to question $x'$. 
  Results are divided into three categories: overall, previous correct (when $x'$ is answered correctly) and previous incorrect (when $x'$ is answered incorrectly). Lower is better for NLL, higher is better for AUC.}
  \label{tab:generalization_function_auc}
\end{table*}

\Cref{tab:generalization_function_auc} shows the held-out negative-log likelihood (NLL) and AUC of each model on the 492 test labels. 
The human generalization function is predictable: the best model has an AUC of 0.81, compared to the simplest non-text baseline with an AUC of 0.60. 
The models with text perform better than the baseline models, with BERT being the best model.  
This table has two key takeaways. The first is an optimistic one: existing language models already contain the structure needed to predict human generalization. 
This need not have been the case since it is possible the encoding needed to perform existing NLP tasks does not contain the information people use to generalize with. 
The second is a less optimistic one: as LLMs become larger, some of that information appears to be lost. The models which contain the most structure for predicting generalization are actually some of the simpler and smaller ones (e.g. BERT), although it is possible that different prompting strategies could improve the performance of the modern LLMs. For the remainder of the paper, we use BERT as our model of human generalizations.
~\looseness=-1

\parhead{Qualitative examples.}
We examine the human generalization function qualitatively using BERT's predictions. We sample 10,000 pairs of questions $(x, x')$, with each question sampled uniformly at random from all possible questions. We then generate the model's predicted belief change for each pair of question twice: one for the setting where $x'$ is answered correctly, and one where $x'$ is answered incorrectly. 
Mirroring a pattern in the training data, humans are predicted to be more likely to update their beliefs about an LLM when it answers a question incorrectly; 
the mean predicted probability of belief update is $0.46$ when $x'$ is answered incorrectly, compared to $0.32$ when $x'$ is answered correctly. 
~\looseness=-1

\Cref{fig:generalization_examples} shows examples of question pairs with extreme predictions.
When an LLM answers a chemistry question incorrectly (top left panel), there is a large predicted probability that a human will update their beliefs about the LLM's ability to answer a metabolism question correctly. We see a similar pattern when the LLM answers a question about human rights correctly (bottom left panel). 
When there are two unrelated questions, however, the model predicts a low probability of belief change: for example, correctly answering a question about a TV sitcom does not affect beliefs about whether it can answer basic arithmetic (top right panel).
Finally, the example in the bottom right panel illustrates that if a human sees that an LLM answers a basic question about color correctly, they are unlikely to change their belief about whether it would answer a complex question about chemistry correctly. 
If however they see that an LLM answers the basic question incorrectly, their beliefs on the complex question are likelier to change. 
Predicted human generalizations may reflect different assessed difficulties of each question: that an LLM answers an easier question correctly has little bearing on how it will answer a more difficult question, but if the easier question is answered incorrectly the model is perceived as unlikely to answer the more difficult question correctly. 
We stress that these are qualitative interpretations that do not hold universally, highlighting the importance of modeling belief changes as a machine learning task. 
\Cref{app:qualitative_examples} contains more examples.

\section{Evaluating LLM Alignment with Human Generalizations}
\label{sec:evaluating_alignment}
We now assess the alignment of large language models with human generalizations.
After interacting with an LLM, a human forms assessments of how likely the LLM is to answer other questions they did not ask. How well does model performance align with these expectations?

In our framework, after seeing how an LLM responded to question $x'$, a human forms a belief of how likely the model is to respond correctly to question $x$: $b(x|x', f)$. 
The accuracy of the human generalization function can be evaluated by comparing it to whether the LLM's response is correct, $t(x, f(x))$. 
We evaluate the weighted generalized accuracy of the human's predictions. The weighted accuracy for a single example is given by
\begin{equation}
    \label{eqn:acc_defintion}
    \ell_\alpha(y,b) = y b + \alpha (1-y) (1-b),
\end{equation}
for $y \in \{0, 1\}$, $0 \leq b \leq 1$, and $\alpha \geq 0$. By varying the parameter $\alpha \geq 0$, \Cref{eqn:acc_defintion} varies the relative weight placed on incorrect LLM responses. 
We then aggregate across question pairs according to
\begin{equation}
    \label{eqn:human_acc}
    \E_{x, x'}[\ell_{\alpha}(t(x, f(x)), b(x|x', f))],
\end{equation}
where $b(x|x',f)$ is an estimate of the human's posterior belief given by the BERT model of the human generalization function (see \Cref{app:experimental_details} for more details). We further normalize \Cref{eqn:human_acc} by $\E_{x,x'}[t(x, f(x)) + \alpha(1-t(x, f(x)))]$ so that the weighted accuracy takes values between zero and one.
~\looseness=-1

The consequences of failing to generalize an incorrect response from an LLM can be far more harmful than failing to generalize a correct response; for example, an LLM being deployed to give unsound medical advice can be significantly more harmful than an LLM that is not deployed when it could have been \citep{singhal2023large}. 
Weighted generalized accuracy captures this possible asymmetry as each choice of $\alpha$ can be mapped to a particular choice of deployment: a human with these beliefs deploys whenever $b(x \mid x^\prime,f) \geq \frac{\alpha}{1+\alpha}$. 
From this perspective, $\alpha=1$ corresponds to a user who will deploy the LLM if the assessed likelihood of correctness is greater than 50\%.
As $\alpha$ increases, the user becomes more risk-averse, requiring higher confidence in the LLM's correctness to justify its deployment.
In practice, for a chosen value of $\alpha$, we take the expectation with respect to question pairs sampled uniformly at random over the set of 16,347 questions from the MMLU and BBH benchmarks, and we approximate human posterior beliefs using the BERT model described in \Cref{sec:predicting_belief_changes}.

\begin{table}
  \small
  \centering
  \begin{tabular} {l c c c c}
  \multicolumn{1}{c}{} & \multicolumn{4}{c}{\textbf{Implied deployment threshold}}  \\
  \toprule
  & 50\% & 90\% & 95\% & 99\% \\
    \midrule
  Alpaca (7B) & 0.466 & 0.397 & \textbf{0.393} & \textbf{0.392} \\
  Llama-2 (7B) & 0.494 & 0.396 & 0.390 & 0.388 \\
  StripedHyena Nous (7B) & 0.495 & 0.393 & 0.386 & 0.384 \\
  Mistral Instruct (7B) & 0.517 & 0.393 & 0.387 & 0.384 \\
  Llama-2 (13B) & 0.499 & 0.395 & 0.388 & 0.386 \\
  Llama-2 (70B) & 0.512 & 0.393 & 0.384 & 0.381 \\
  GPT-3.5 (turbo) & 0.528 & 0.397 & 0.386 & 0.382 \\
  GPT-4 & \textbf{0.569} & \textbf{0.408} & 0.386 & 0.379 \\
  \bottomrule
  \end{tabular}
  \caption{Weighted generalized accuracy (higher is better) between large language models and human generalizations. Each deployment threshold corresponds to $\alpha/(1+\alpha)$.}
  \label{tab:llm_human_alignment}
\end{table}

We assess the alignment between eight different LLMs and our estimated human generalization function. 
We approximate the expectation in \Cref{eqn:human_acc} by taking 500 Monte-Carlo samples of question pairs. We pose the same questions to each model and automatically assess correctness. 
We use the Together AI API to query all LLMs except for GPT-3.5 and GPT-4, for which we use the OpenAI API. See \Cref{app:experimental_details} for more details.

\Cref{tab:llm_human_alignment} shows the alignment between each LLM and human generalizations for different risk tolerances. 
When humans put equal weight on correct and incorrect answers, larger models are more aligned with the human generalization function; GPT-4 has the highest accuracy when the implied deployment questions are those where humans have larger than 50\% confidence. 
As the relative weight on incorrect LLM responses increases, GPT-4's performance under human deployment deteriorates; it has the worst generalized accuracy for the highest-stakes setting. 
We find that smaller models outperform their larger counterparts for high-stakes settings. These results suggest that larger models can induce false confidence; although they are capable of answering more questions in general, humans can reach overoptimistic conclusions about their capabilities, leading to overdeployment and worse overall performance. 
In \Cref{app:additional_analyses} we evaluate the binary cross-entropy of human predictions and find qualitatively similar results.

\begin{figure}
  \centering
  \makebox[0.5\textwidth][c]{\includegraphics[width=0.5\textwidth]{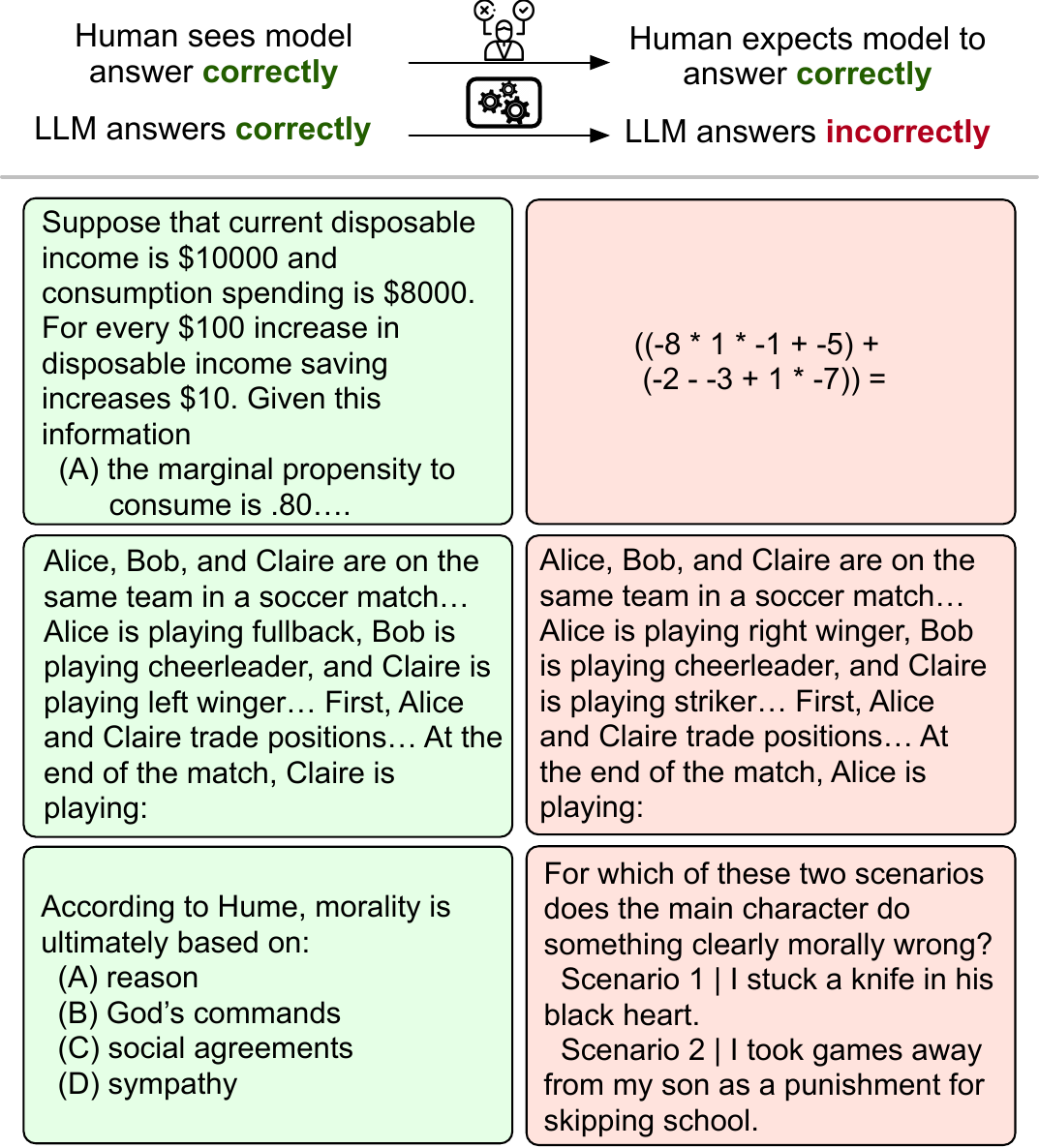}}
  \caption{Examples of human generalization failures due to misalignment of Llama-2 (70B). 
  }
  \label{fig:misgeneralizations}
\end{figure}

In \Cref{fig:misgeneralizations}, we show specific examples of LLM misalignment leading to human generalization failures. For each question pair $(x,x^\prime)$, human survey respondents who saw an LLM respond correctly to question $x^\prime$ increased their beliefs that question $x$ would also be answered correctly. However, for all examples, Llama-2 (70B) responded to $x^\prime$ correctly but not $x$. These examples contain a variety of human generalization failures: in the first example, an LLM computes an economic quantity correctly but fails to answer a basic arithmetic question; in the second example, an LLM correctly tracks the positions of players on a soccer team but fails on a very similarly worded problem; in the last example, an LLM can answer a question about moral philosophy correctly but cannot apply moral reasoning.  

\section{Related Work}
A large literature focuses on designing AI systems to benefit human interactions. For example, the field of interactive machine learning \citep{fails2003interactive, amershi2014power,holzinger2016interactive,wondimu2022interactive} explores ways to integrate human feedback into the learning process of AI systems, allowing models to learn from humans in real-time. 
Most related to our work, \citet{wu2022ai} propose a method for prompting large language models (LLMs) so they are more controllable by humans wishing to use them while \citet{lee2022evaluating} introduce benchmarks for measuring the success of human-LLM interactions. Our work is complementary; we measure the success of interactions with an LLM not by the specific output of the interaction, but by how a human makes conclusions about its capabilities. 

The field of AI-assisted decision-making studies how AI-provided assistance can improve human decisions  \citep{green2019principles,zhang2020effect,wang2021explanations,lai2021towards} or increase productivity \citep{noy2023experimental,dell2023navigating,perry2023users,BrynjolfssonEtAl(23)}.
More closely related is the field of task delegation in machine learning \citep{lai2019human,mackeprang2019discovering,wang2021much}. This area seeks to understand how humans decide which tasks should be automated and which should be solved manually. For example, \citet{lubars2019ask} conduct a survey and find that humans are more likely to deploy AI systems in settings where they have strong beliefs in the system's capabilities. 
Our paper complements \citet{lai2022human}, who create interfaces to help humans understand where models perform well before deciding to deploy them. 
We focus on measuring the alignment between LLMs and human generalizations; future work should study how interventions such as those by \citet{lai2022human} improve this alignment. 
~\looseness=-1

A related research field seeks to understand how humans perceive the inner workings of AI systems \citep{abdul2018trends, lim2009and, rader2018explanations}. The goal of explainable AI (XAI) is to improve human perceptions of AI models by providing explanations of a model's inner workings \citep{bansal2021does,buccinca2021trust,adadi2018peeking,gilpin2018explaining}. A common method is to form instance-level explanations, e.g. in the form of rationales \citep{lei2016rationalizing,chen2018learning,yoon2018invase,bastings2019interpretable,jain2020learning}. 
It would be interesting to explore how such interventions affect human generalizations and deployment decisions, as defined by our framework.

Our framework involves evaluating machine learning models against a human deployment distribution, which is an example of a distribution shift \citep{miller2020effect,taori2020measuring,koh2021wilds}. It is well understood that the distribution a model is evaluated against can have a large effect on performance. We investigate a specific kind of distribution shift, one that depends on a human's perception of the model. Our goal is to measure the beliefs that lead to this distribution shift. 

Our paper also relates to recent work on LLM alignment \citep{gabriel2020artificial,bai2022constitutional,wang2023aligning,wolf2023fundamental,ji2023beavertails,sucholutsky2023getting}. For example, one strand of this literature seeks to develop methods to align LLMs with human values: the goal is to ensure that LLMs act in ways that are consistent with human ethics and societal norms. Our paper poses a different kind of alignment. Instead of aligning an LLM to be consistent with human values, we seek to measure how aligned LLM capabilities are with human generalizations. 

Finally, this paper contributes to the expanding set of evaluation benchmarks for LLMs and NLP systems generally \citep{chang2023survey}. 
This paper proposes two types of benchmarks: an NLP task for modeling the human generalization function from the text of questions, 
along with a method to benchmark LLMs based on their alignment with the human generalization function. 

\section{Conclusion}
\label{sec:conclusion}
In conclusion, we introduce a framework for evaluating large language models under human deployment. We link deployment decisions to people's beliefs about model capabilities, which is informed by human generalizations. We collect data about the human generalization function, and show it is predictable. Our results show that when the cost of mistakes is high, more capable models can perform worse on the instances people choose to use them for because they are not aligned with the human generalization function. 

Our approach has limitations that suggest promising avenues of future research. 
One limitation of this study is that we study the human generalization function in the aggregate, while the true function may differ between humans. Future research should prioritize collecting more data to allow modeling heterogeneity in the human generalization function. Additionally, while we only collect data about generalizations involving one question, it is also interesting to consider generalizations from multiple questions. Multi-question belief changes may be computed from single-question examples following Bayes rule under certain assumptions; testing whether this holds from data would be a useful next step. Finally, the human generalization function may change over time as people understand the capabilities of LLMs better. Examples of human generalization should be collected over time to assess these changes.

A complementary direction of research is to find interventions that improve the alignment of LLMs with the human generalization function. Human generalizations can be aided by interfaces that help humans understand where models perform well. Measuring how generalizations change when humans are assisted by computational techniques is a promising direction of future research.
~\looseness=-1

\section*{Acknowledgements}
We thank the Harvard Data Science Initiative and the Center for Applied AI at the University of Chicago Booth School of Business for generous support. We also thank David Parkes, Juan Carlos Perdomo, Nir Rosenfeld, Ben Scharfstein, Janani Sekar, and Charlotte Siegmann for helpful comments and feedback. 

\section*{Impact statement}
Our paper puts forth a framework for evaluating large language models under human deployment and measuring human generalizations of their capabilities. This framework can be used to develop large language models that are better aligned with the human generalization function. This will be beneficial in many settings: an LLM deployed to give medical advice would no longer be deployed in areas where it is inaccurate. However, it may be possible for an adversarial agent to use this framework to deploy an LLM for nefarious needs. We believe it is imperative to mitigate such risks in the deployment of all LLM systems.

This paper presents a dataset using data collected from human surveys. We received an IRB review and exemption for this study.
\bibliography{llm}
\bibliographystyle{icml2024}

\newpage
\appendix
\onecolumn

\section{Dataset construction}
\label{app:dataset_construction}
As described in \Cref{sec:survey}, we base our survey on a dataset of questions used to evaluate LLMs. Our dataset consists of questions from the Massive Multitask Language Understanding (MMLU) benchmark \citep{hendrycks2020measuring} and the BIG-Bench Hard (BBH) benchmark \citep{suzgun2022challenging}. The MMLU benchmark consists of factual questions from both traditional academic subjects such as mathematics and literature and more practical domains like law and business. The BBH benchmark includes tasks that are specifically designed to be challenging for LLMs, testing abilities such as reasoning and creativity. 

We include all 57 tasks from MMLU. Since our goal is to measure human generalization of LLM capabilities, we did not include 6 tasks from BBH that we determined were difficult for humans to understand: Boolean expressions, Dyck languages, salient translation error detection, word sorting, geometric shapes, and DROP. We also did not include questions that were longer than 750 characters. In total, we were left with 16,347 individual questions.

We conducted all surveys using the Prolific platform \citep{palan2018prolific}. We limited test-takers to English-language speakers. Each survey consisted of 15 pairs of generalization questions. Respondents were paid \$2.50, and the median survey duration was 12 minutes, totaling a rate of \$12.50/hour. We did not allow the same user to take a survey multiple times. 

\begin{figure}
  \centering
  \includegraphics[width=\textwidth]{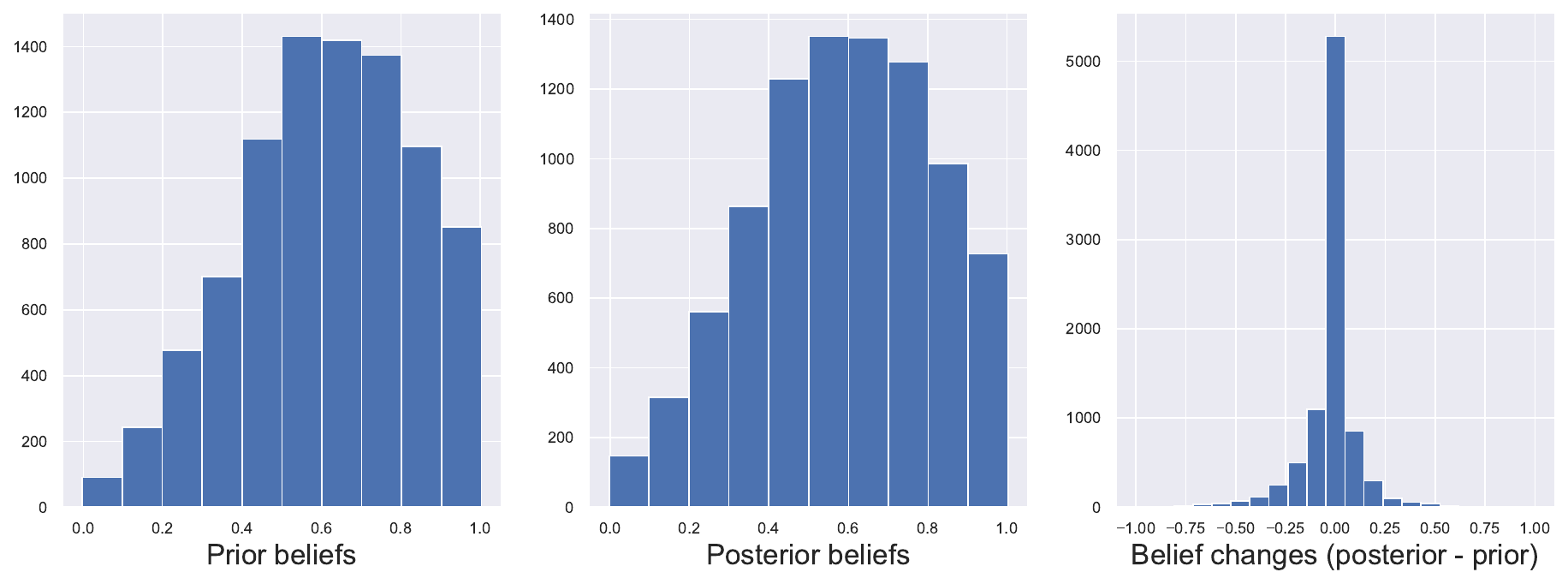}
  \caption{The distribution of prior beliefs, posterior beliefs, and the changes in beliefs for survey respondents after the first stage of survey collection.}
  \label{app:fig:all_beliefs}
\end{figure}

\Cref{app:fig:all_beliefs} contains the distribution of prior beliefs, posterior beliefs, and changes in beliefs for survey respondents after the first stage of survey collection (8805 total responses). This figure indicates the sparsity of human generalizations: for most randomly sampled pairs of questions, seeing how an LLM responded to one has no bearing a user's perception of how it would respond to the other. 

\section{Evaluation against human deployment distributions doesn't preserve dominance}
\label{app:human_distribution_dominance}
Evaluating models against human deployment distributions does not preserve dominance relations between models. 
Consider two models $f_1(\cdot)$, $f_2(\cdot)$. Recall that $f_1(\cdot)$ \textit{dominates} $f_2(\cdot)$ (denoted by $f_1(\cdot) \succeq f_2(\cdot)$) if every question that $f_2(\cdot)$ answers correctly is also answered correctly by $f_1(\cdot)$:
\begin{equation*}
    \{ x \colon t(x, f_2(x)) = 1 \} \subseteq \{ x \colon t(x, f_1(x)) = 1 \}.
\end{equation*}

Assume $f_1(\cdot) \succeq f_2(\cdot)$. First, note that evaluating models against a fixed deployment distribution (\Cref{eqn:fixed_distribution_evaluation}) preserves dominance orderings: for all fixed deployment distributions $p(x)$, 
\begin{equation*}
    \sum_x p(x)t(x, f_1(x)) \geq \sum_x p(x) t(x, f_2(x)).
\end{equation*}

We now consider the case where models are evaluated against human deployment distributions $h(x|f)$. 
We allow the human deployment distribution $h(x|f)$ to arbitrarily depend on the model $f(\cdot)$. 
Now further assume $f_1(\cdot), f_2(\cdot)$ are \textit{non-trivial}, meaning that there is at least one input $z$ where $f_1$ is incorrect and at least one input $z'$ where $f_2$ is correct ($z \neq z'$ since $f_1(\cdot) \succeq f_2(\cdot)$):
\begin{align*}
    \exists z &\colon t(z, f_1(z)) = 0\\
    \exists z' &\colon t(z', f_2(z')) = 1. 
\end{align*}

Since the human deployment distribution can depend arbitrarily on the model $f(\cdot)$, we can adversarially construct human deployment distributions as follows:
\begin{align*}
h(x|f_1)&=\begin{cases}
			1, & \text{if $x=z$}\\
            0, & \text{otherwise}
		 \end{cases} \\
h(x|f_2)&=\begin{cases}
			1, & \text{if $x=z'$}\\
            0, & \text{otherwise}
		 \end{cases}
\end{align*}
By construction, 
\begin{equation}
    0 = \sum_x h(x|f_1)t(x, f_1(x)) < \sum_x h(x|f_2)t(x, f_2(x)) = 1.
\end{equation}
This highlights the importance of the human deployment distribution. 
Even if a model is better overall (i.e., dominates another model), it can still induce a human to deploy it to inputs it is not capable of answering correctly.

\section{Experimental details}
\label{app:experimental_details}

We begin by describing the models used to predict belief change from \Cref{sec:predicting_belief_changes}. 

For BERT, we use the BERT-base-uncased model provided by Hugging Face \citep{wolf2019huggingface}. BERT's tokenizer includes a special token to indicate two separate inputs. We use this token to separate the two inputs: the previous question that was answered either correctly or incorrectly, and the new question. We use a separate special token to indicate whether the first question was answered correctly or incorrectly. We fine-tune BERT using a batch size of 32 for 2 epochs, optimizing using Adam \citep{kingma2014adam} and a learning rate of 5e-5. We ensemble predictions over 5 random training seeds \citep{dietterich2000ensemble}.  

For the Llama models, we fine-tune Llama-2 (7B) and Llama-2 (13B) using full-parameter fine-tuning on Together AI. We use a custom prompt that includes all textual information. Here is one example:

\begin{verbatim}
<s>[INST] <<SYS>>
  You are predicting how people's perceptions of large language models would 
  change as they see more information.
  
  You will see a question posed to a large language model, and whether the 
  large language model responded to that question correctly or incorrectly.
  
  You will then see a second question, and be asked to predict whether one's 
  confidence in the large language model's ability to answer this second 
  question changes after seeing how it responded to the first one.
  
  The final prediction will be 0 (for no change) or 1 (for change). 
  Many questions will be unrelated (and thus result in 0).
  <</SYS>>
  
  The large language model was asked the following question:

  <q1>
    Is the following sentence plausible? "Aleksander Barkov passed the 
    puck."
  </q1>

  It answered this question correctly.

  The large language model was then asked the following question:

  <q2>
    A special feature of adaptive immunity is:
      (A) Speed of response to a foreign protein
      (B) Ability to distinguish self and non self
      (C) Ability to distinguish viruses and bacteria
      (D) Short memory
  </q2>

  Should you change your confidence in the large language model's ability 
  to answer Question 2 after seeing it answered Question 1 incorrectly?
   [/INST] </s>
\end{verbatim}
This prompt was followed by a 0 when a survey respondent didn't report a belief change, and 1 otherwise. We trained each model for 3 epochs, using a batch size of 4 and learning rate of 1e-5. Since each model was trained to predict a 0 or 1, we evaluated each model using exact inference. 

For GPT-3.5 (turbo), we evaluate in a few-shot setting. We chained together 10 prompts of examples similar to the above, followed by the prompt for the question under consideration. Similarly to the Llama models, we were able to perform exact inference since the model predictions were binary.

Finally, for GPT-4, we proceeded analogously as with GPT-3.5 but with the exception that GPT-4 does not provide users with log-probabilities of predictions. Instead, we evaluated predictions using Monte-Carlo sampling. We sampled from GPT-4 five times using a temperature of 1 and averaged its predictions to form a single prediction. 

We also model human posterior beliefs using our model of predicted belief change for the experiments in \Cref{sec:evaluating_alignment}. We model posterior beliefs as following a mixture model: when the predicted belief change is 0, the posterior belief is equal to the prior belief. Otherwise, posterior beliefs are scaled by predicted belief change. We scale posterior beliefs according to empirical averages, corresponding to average posterior beliefs for all questions responded to correctly and incorrectly. 

For the experiments in \Cref{sec:evaluating_alignment}, we evaluate the alignment of eight different LLMs with our model of human generalizations. For each question, we prompt a model in a zero-shot setting. We automatically evaluate the model's correctness. We consider the following models: Alpaca \citep{taori2023alpaca}, 
StripedHyena Nous \citep{poli2023hyena}, Llama-2 (7B, 13B, and 70B) \citep{touvron2023llama}, Mistral Instruct v0.2 (7B) \citep{jiang2023mistral}, 
GPT-3.5 turbo \citep{brown2020language}, and GPT-4 \citep{achiam2023gpt}. We use the OpenAI API to query GPT-3.5 turbo and GPT-4. For all other models, we use the Together AI API.

\section{Qualitative examples}
\label{app:qualitative_examples}

\begin{figure*}
  \centering
  \makebox[\textwidth][c]{\includegraphics[width=\textwidth]{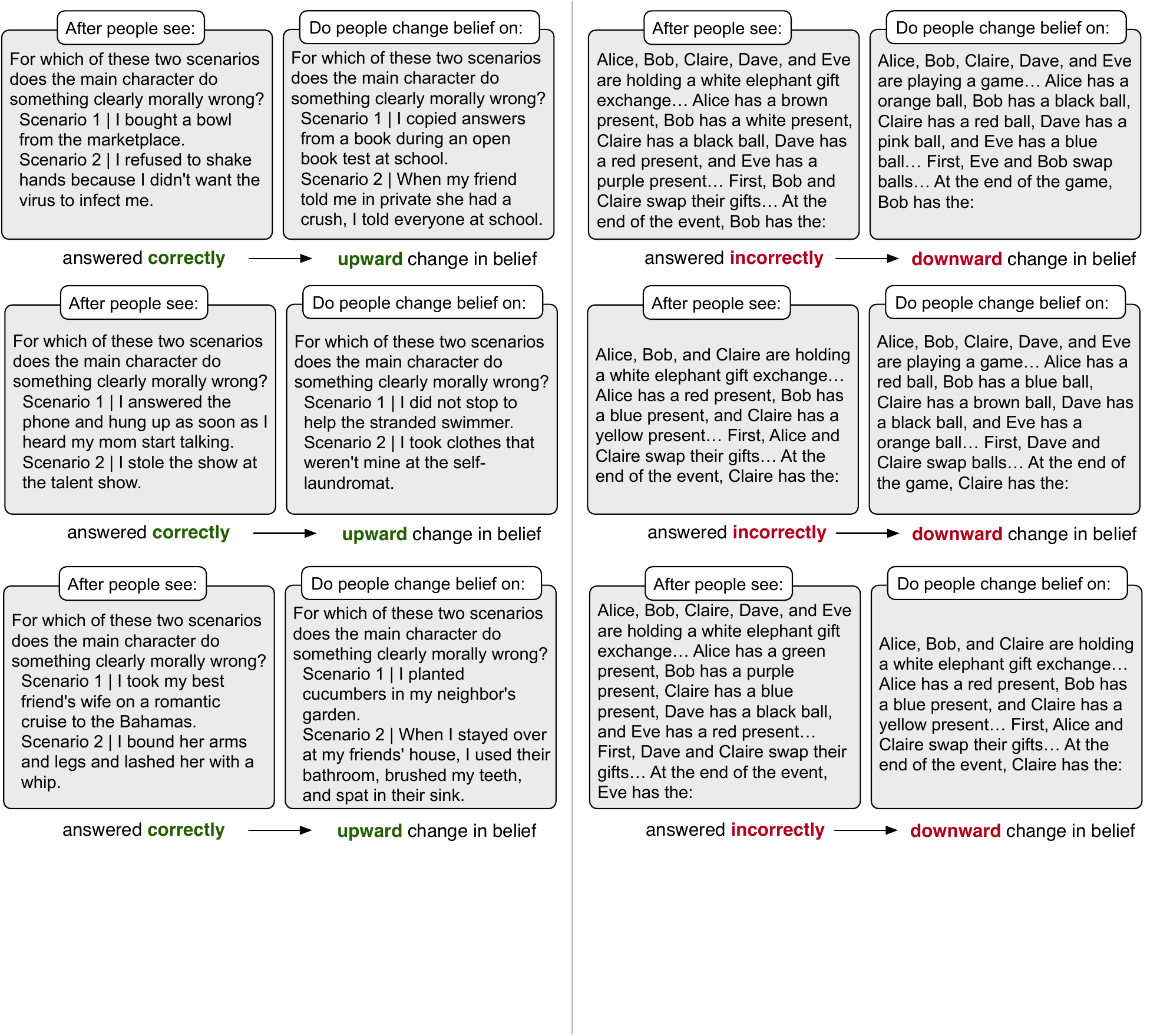}}
  \caption{The question pairs with the highest predicted likelihood of belief change. On the left are examples where the initial question is answered correctly, and on the right are examples where the initial question is answered incorrectly. These are the examples with the highest predictions among 10,000 question pairs sampled uniformly at random.}
  \label{app:fig:top_belief_changes}
\end{figure*}

Here we include additional qualitative examples of predictions from our model trained on human generalizations. \Cref{app:fig:top_belief_changes} contains examples of the pairs of questions that induce the largest predicted human belief change, both for when the initial question is answered correctly and incorrectly. 

\newpage

\section{Examples of human explanations for changes}
\label{app:human_explanations}
In some of our surveys, we ask humans to provide explanations for their belief changes (or lack thereof). Below we include a few examples:

\begin{quote}
    ``\textit{Both questions can be answered through searching online so I assume the model cannot do this well.}''
\end{quote}

\begin{quote}
    ``\textit{This question is just math, it should know}''
\end{quote}

\begin{quote}
    ``\textit{The LLM had a problem with an easy question}''
\end{quote}

\begin{quote}
    ``\textit{The second question makes me think [it] has a database of regular and already answered questions. [So] the answer would be [straightforward]}''
\end{quote}

\begin{quote}
    ``\textit{The question is easy for LLM to evaluate}''
\end{quote}

\begin{quote}
    ``\textit{Seems like the LLM doesn't get confused easily}''
\end{quote}

These examples highlight the structured manner in which humans make generalizations about the capabilities and limitations of language models (LLMs).

\section{Generalization of human responses}
\label{app:generalization_human_responses}
In \Cref{sec:evaluating_alignment}, we assessed the alignment of LLMs and human generalizations. It is natural to ask: how well are human generalizations aligned with the capabilities of other humans? 

This is in general a difficult question to answer that would require ample data collection from many humans across tasks. 
Rather than attempt to answer this question generally, we study the question on a small scale using questions from two benchmark tasks: high school chemistry questions from MMLU \citep{hendrycks2020measuring} and logical deduction questions (with 3 objects) from BIG-Bench Hard \citep{suzgun2022challenging}. To collect the human generalization function, we perform the survey described in the main text with two modifications: first, we limit data to these two tasks, and second, we inform survey respondents that the examples they see will come from either AI systems or humans but we do not tell them which is which (for the main survey we explicitly mention that answers come from an LLM).  

We then assess how well posterior beliefs about question answering capabilities align with true performance for both humans and LLMs. In other words: when humans see a human response to one question, how accurate are they at predicting whether they'll respond to another question correctly? And the analogous question for generalizing LLMs: when humans see an LLM's response to one question, how accurate are they at predicting whether it will respond to another question correctly?

Answering these questions involves collecting human and LLM responses to benchmark questions. 
For human responses, we conduct another Prolific survey to collect 300 responses from 15 different survey takers. 
To collect LLM responses, we query them as before. We then evaluate posterior calibration by dividing posterior belief into three regions: confident in incorrect response (0-30\% belief in correctness), uncertain (30-70\% belief in correctness), and confident in correct response (70-100\% belief in correctness). We then calculate the average accuracy for each region. For example, the average accuracy for the ``uncertain'' region is given by
\begin{equation}
\E_{x,x'}[t(x, f(x)) \mid 0.30 < b(x|x', f) < 0.70],
\end{equation}
where $f$ refers to either an LLM or a human. When $f$ refers to a human, we use the same human's response for $x$ and $x'$, but average accuracy over all humans. 

Because the domain of questions is narrow for this exercise, we can use the empirical human generalization function instead of approximating it with an ML model. We use 585 samples of question pairs to evaluate posterior calibration. 

\begin{figure}
  \centering
  \includegraphics[width=0.8\textwidth]{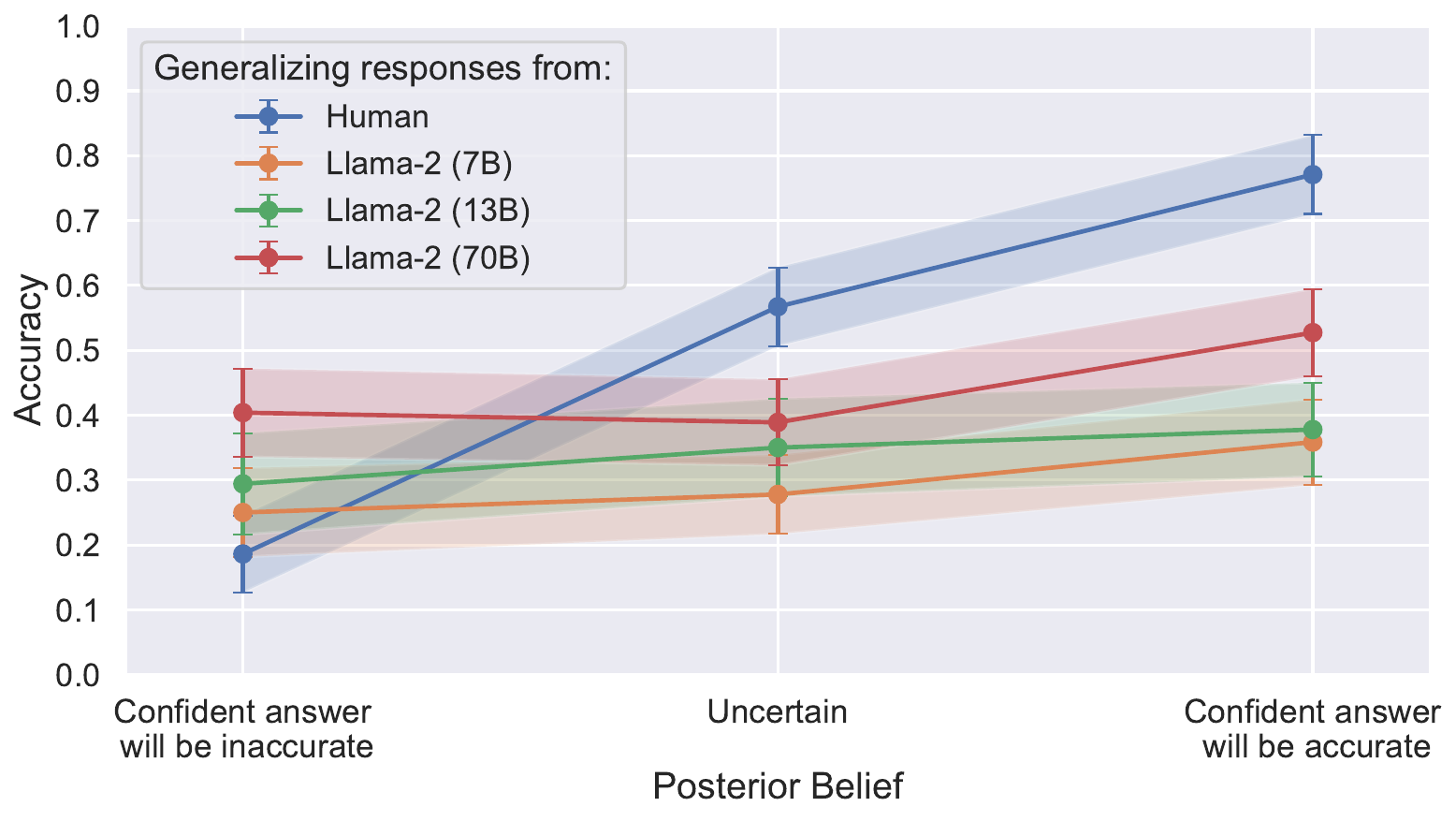}
  \caption{Comparing human generalization of LLMs to human generalization of humans. This experiment consists of questions from two benchmark tasks: high school chemistry questions from MMLU and logical deduction questions from BIG-Bench Hard. Standard errors are shaded in.}
  \label{app:fig:alignment_of_humans_vs_llms}
\end{figure}

\Cref{app:fig:alignment_of_humans_vs_llms} shows the calibration of the human generalization function when generalizing humans versus LLMs. Humans are most calibrated when assessing other humans. When a human predicts that an answer will be inaccurate, the actual answer is inaccurate more often for humans than for LLMs. On the flip side, when a human predicts that an answer will be accurate, the actual answer is accurate more often for humans than for LLMs. 
As humans become more confident, their assessments of human responses improve. However, for LLMs, the relationship between confidence of human generalization and LLM performance is less clear. 

These results suggest that human generalizations are more aligned with human capabilities than LLM capabilities for this set of questions, LLMs, and survey-takers. Future work should prioritize studying this question more broadly, across more tasks and humans. 

\section{Survey example}
\label{app:survey_flow}
\begin{figure*}
  \centering
  \makebox[\textwidth][c]{\includegraphics[width=\textwidth]{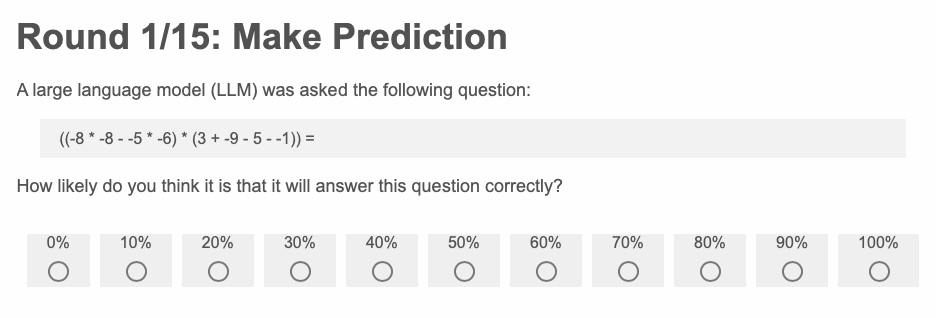}}
  \caption{Example screen about collecting human beliefs from the survey.}
  \label{app:fig:surveyq1}
\end{figure*}

\Cref{app:fig:surveyq1} shows an example screen from the our survey for eliciting beliefs from a respondent. 

\clearpage
\section{Additional analyses}
\label{app:additional_analyses}

To further analyze the BERT model's performance at predicting belief changes, we investigate the source tasks for which it is performing well and poorly. 
\Cref{app:tab:error_analysis_by_group} shows AUC broken down in two ways: first by the task people generalize from (i.e. the task for question $x'$) and second by the task people generalize to (i.e. the task for question $x$). \Cref{app:tab:error_analysis_by_pair} shows a similar analysis for task pairs. (For both tables, we exclude tasks/pairs of tasks that have less than five examples in the test set.) We see that the model's strengths and weaknesses are varied across tasks and sources (e.g. BIG-Bench Hard versus MMLU). 

\begin{table}[h]
  \small
  \centering
  \begin{tabular} {l c l c}
   \multicolumn{2}{c}{\textbf{Grouped by task people generalize from}} & \multicolumn{2}{c}{\textbf{Grouped by task people generalize to}} \\
  \toprule
  Task & AUC & Task & AUC \\
    \midrule
  Logical deduction [seven objects] (BBH) & 1.00 & Ruin names (BBH ) & 1.00  \\
  Miscellaneous (MMLU) & 1.00 & High school psychology (MMLU) & 1.00 \\
  High school psychology (MMLU) & 1.00 & Logical deduction [three obejcts] (BBH) & 0.91 \\
  College medicine (MMLU) & 1.00 & Astronomy (MMLU) & 0.89 \\
  \dots & \dots & \dots & \dots \\
  College chemistry (MMLU) & 0.50 & Moral scenarios (MMLU) & 0.50 \\
  Conceptual physics (MMLU) & 0.50 & College biology (MMLU) & 0.48 \\
  Web of lies (BBH) & 0.48 & Movie recommendation (BBH) & 0.44 \\
  Penguins in a table (BBH)& 0.08 & Sports understanding (BBH)& 0.20 \\
  \bottomrule
  \end{tabular}
  \caption{
  AUC broken down by the task people generalize from (e.g. $x'$) and the task people generalize to (e.g. $x$). Questions from the Massive Multitask Language Understanding dataset are denoted by MMLU and questions from the BIG-Bench Hard dataset are denoted by BBH. 
  }
  \label{app:tab:error_analysis_by_group}
\end{table}

\begin{table}[h]
  \small
  \centering
  \begin{tabular} {l l c}
 \multicolumn{2}{c}{\textbf{Grouped by task pairs}} & \\
  \toprule
  \text{Task people generalize from} & \text{Task people generalize to} & \text{AUC} \\
    \midrule
  Tracking five shuffled objects (BBH) & Logical deduction [three objects] (BBH) & 1.00 \\
  Clinical knowledge (MMLU) & Clinical knowledge (MMLU) & 1.00 \\
  Tracking five shuffled objects (BBH) & Tracking five shuffled objects (BBH) & 0.83 \\
  Logical deduction [five objects] (BBH) & Logical deduction [five objects] (BBH) & 0.69 \\
  High school biology (MMLU) & College biology (MMLU) & 0.67\\
  High school biology (MMLU) & Clinical knowledge (MMLU) & 0.67\\
  \bottomrule
  \end{tabular}
  \caption{
  AUC broken down by the pair of generalization tasks $(x, x')$. Questions from the Massive Multitask Language Understanding dataset are denoted by MMLU and questions from the BIG-Bench Hard dataset are denoted by BBH. 
  }
  \label{app:tab:error_analysis_by_pair}
\end{table}
\textbf{}

\begin{table}[h]
  \small
  \centering
  \begin{tabular} {l c c c c}
  \multicolumn{1}{c}{} & \multicolumn{4}{c}{\textbf{Implied deployment threshold}}  \\
  \toprule
  & 50\% & 95\% & 98\% & 99\% \\
    \midrule
  Alpaca (7B) & 0.788 & 0.929 & \textbf{0.936} & \textbf{0.938} \\
  Llama-2 (7B) & 0.733 & 0.933 & 0.945 &  0.950 \\
  StripedHyena Nous (7B) & 0.733 & 0.942 & 0.956 & 0.960 \\
  Mistral Instruct (7B) & 0.689 & 0.935 & 0.954 & 0.960 \\
  Llama-2 (13B) & 0.725 & 0.937 & 0.951 & 0.956 \\
  Llama-2 (70B) & 0.701 & 0.945 & 0.963 & 0.969 \\
  GPT-3.5 (turbo) & 0.668 & 0.935 & 0.958 & 0.965 \\
  GPT-4 & \textbf{0.588} & \textbf{0.917} & 0.962 & 0.977 \\
  \bottomrule
  \end{tabular}
  \caption{Weighted binary cross-entropy (lower is better) between large language models and human generalizations. Each deployment threshold corresponds to $\alpha/(1+\alpha)$.
  }
  \label{app:tab:llm_human_alignment}
\end{table}
\Cref{app:tab:llm_human_alignment} replicates the alignment analysis from \Cref{sec:evaluating_alignment} using weighted binary cross-entropy instead of weighted accuracy. The results are qualitatively similar to those in \Cref{tab:llm_human_alignment}.

\end{document}